\documentclass{IEEEtran}
\usepackage{makeidx}
\usepackage{algorithm}
\usepackage{algorithmic}
\usepackage{pdflscape}
\usepackage{authblk}
\usepackage{multirow,longtable}
\usepackage{amssymb}
\usepackage{url}
\usepackage[utf8]{inputenc}
\usepackage[T1]{fontenc}
\usepackage{tikz}
\usepackage{amsmath,mathtools}
\usepackage{stmaryrd}
\usepackage{caption}
\usepackage{subfig}
\usepackage{rotating}
\usepackage{blindtext}
\usetikzlibrary{shapes.geometric, arrows}

\title{Gene Similarity-based Approaches\\
for Determining Core-Genes of Chloroplasts}
\author{
\IEEEauthorblockN{Bassam AlKindy\IEEEauthorrefmark{1}\IEEEauthorrefmark{2}, Christophe Guyeux\IEEEauthorrefmark{1}, Jean-Fran\c{c}ois Couchot\IEEEauthorrefmark{1}, Michel Salomon\IEEEauthorrefmark{1}, Jacques M. Bahi\IEEEauthorrefmark{1}}\\
\IEEEauthorblockA{\IEEEauthorrefmark{1}FEMTO-ST Institute, UMR 6174 CNRS, DISC Computer Science Department, \\ University of Franche-Comt\'{e}, France}\\
\IEEEauthorblockA{\IEEEauthorrefmark{2}Department of Computer Science, University of Mustansiriyah, Baghdad, Iraq}\\
\IEEEauthorblockA{\IEEEauthorrefmark{0}\{bassam.al-kindy, christophe.guyeux, jean-francois.couchot,	michel.salomon, jacques.bahi\}@univ-fcomte.fr}
}

\begin{document}
\maketitle

\begin{abstract}\footnote{This paper was submitted to IEEE International Conference on Bioinformatics and Biomedicine (BIBM 2014) on 15/07/2014, Accepted 05/09/2014. Oral presentation was on 05/11/2014.}
In computational biology and bioinformatics, the manner to understand evolution processes within various related organisms paid a lot of attention these last decades. However, accurate methodologies are still needed to discover genes content evolution. In a previous work, two novel approaches based on sequence similarities and genes features have been proposed. 
More precisely, we proposed to use genes names, sequence similarities, or both, insured either from NCBI or from DOGMA annotation tools. Dogma has the advantage to be an up-to-date accurate automatic tool specifically designed for chloroplasts, whereas NCBI possesses high quality human curated genes (together with wrongly annotated ones). 
The key idea of the former proposal was to take the best from these two tools. However, the first proposal was 
limited by name variations and spelling errors on the NCBI side, leading to core trees of low quality.
In this paper, these flaws are fixed by improving the comparison of NCBI and DOGMA results, and
by relaxing constraints on gene names while adding a stage of post-validation on gene sequences.
The two stages of similarity measures, on names and sequences, are thus proposed for sequence clustering. 
This improves results that can be obtained using either NCBI or DOGMA alone. Results obtained
with this ``quality control test'' are further investigated and compared with previously released ones, on both
computational and biological aspects, considering a set of 99 chloroplastic genomes. 

\end{abstract}


\section{Introduction}\label{sec:intro}
The idea motivating the importance of identifying core genes is to understand the shared functionality of a given set of species. 
We introduced in a previous work~\cite{Alkindy2014} two methods for discovering core and pan genes of chloroplastic genomes using both sequence similarity and alignment based approaches. To
determine these core and pan genomes for a large set of DNA sequences, we propose in this work to improve the alignment based approach by considering a novel sequence quality control test. More precisely, we focus on the following questions considering a collection of 99~chloroplasts: how can  we identify the  best core  genome (an artificially designed set of coding sequences as close as possible to the real biological one) and how to deduce scenarii regarding their gene loss.

The term Chloroplast comes from the combination of plastid and chloro, meaning that it is an organelle found in plant and eukaryotic algae cells which contains chlorophyll. Chloroplasts may have evolved from \emph{Cyanobacteria} through endosymbiosis and since their main objective is to conduct photosynthesis, these fundamental tiny energy factories are present in many organisms.  
This key role explains why chloroplasts are at the basis of most trophic pyramids and thus responsible for evolution and speciation. 
Moreover, as photosynthetic organisms release atmospheric oxygen when converting light energy into chemical energy and simultaneously produce organic molecules from carbon dioxide, they originated the breathable air and represent a mid to long term carbon storage medium. Consequently, exploring the evolutionary history of chloroplasts is of great interest and therefore further phylogenetic studies are needed. 

An early study of finding the common genes in chloroplasts was realized in 1998 by \emph{Stoebe et al.}~\cite{stoebe1998distribution}. They established the distribution of 190 identified genes and 66 hypothetical protein-coding genes (\emph{ysf}) in all nine photosynthetic algal plastid genomes available (excluding non photosynthetic \emph{Astasia tonga}) from the last update of plastid genes nomenclature and distribution. The distribution reveals a set of approximately 50 core protein-coding genes retained in all taxa. In 2003, \emph{Grzebyk et al.}~\cite{grzebyk2003mesozoic}, have studied the core genes among 24 chloroplastic sequences extracted from public databases, 10 of them being algae plastid genomes. They broadly clustered the 50 genes from \emph{Stoebe et al.} into three major functional domains: (1) genes encoded for ATP synthesis (\emph{atp} genes); (2) genes encoded for photosynthetic processes (\emph{psa} and \emph{psb} genes); and (3) housekeeping genes that include the plastid ribosomal proteins (\emph{rpl} and \emph{rps} genes). The study shows that all plastid genomes were rich in housekeeping genes with one \emph{rbcLg}  gene involved in photosynthesis.

To determine core chloroplast genomes for a given set of photosynthetic organisms, bioinformatics investigations using sequence annotation and comparison tools are required, and therefore various choices are possible. The purpose of our research work is precisely to study the impact of these choices on the obtained results. 
A general presentation of the approaches we propose is provided in Section~\ref{sec:general}. A closer examination of the approaches is given in Section~\ref{sec:extraction}. Section~\ref{sec:simil} will present coding sequences clustering method based on sequence similarity, while Section~\ref{sec:mixed} will describe quality test method based on quality genes. 
The paper ends with a discussion based on biological aspects regarding the evolutionary history of the considered genomes, leading to our methodology proposal for core and pan genomes discovery of chloroplasts, followed by a conclusion section summarizing our investigations.

\section{An overview of the Pipeline}\label{sec:general}

\begin{figure}[ht]  
  \centering
    \includegraphics[width=0.37\textwidth]{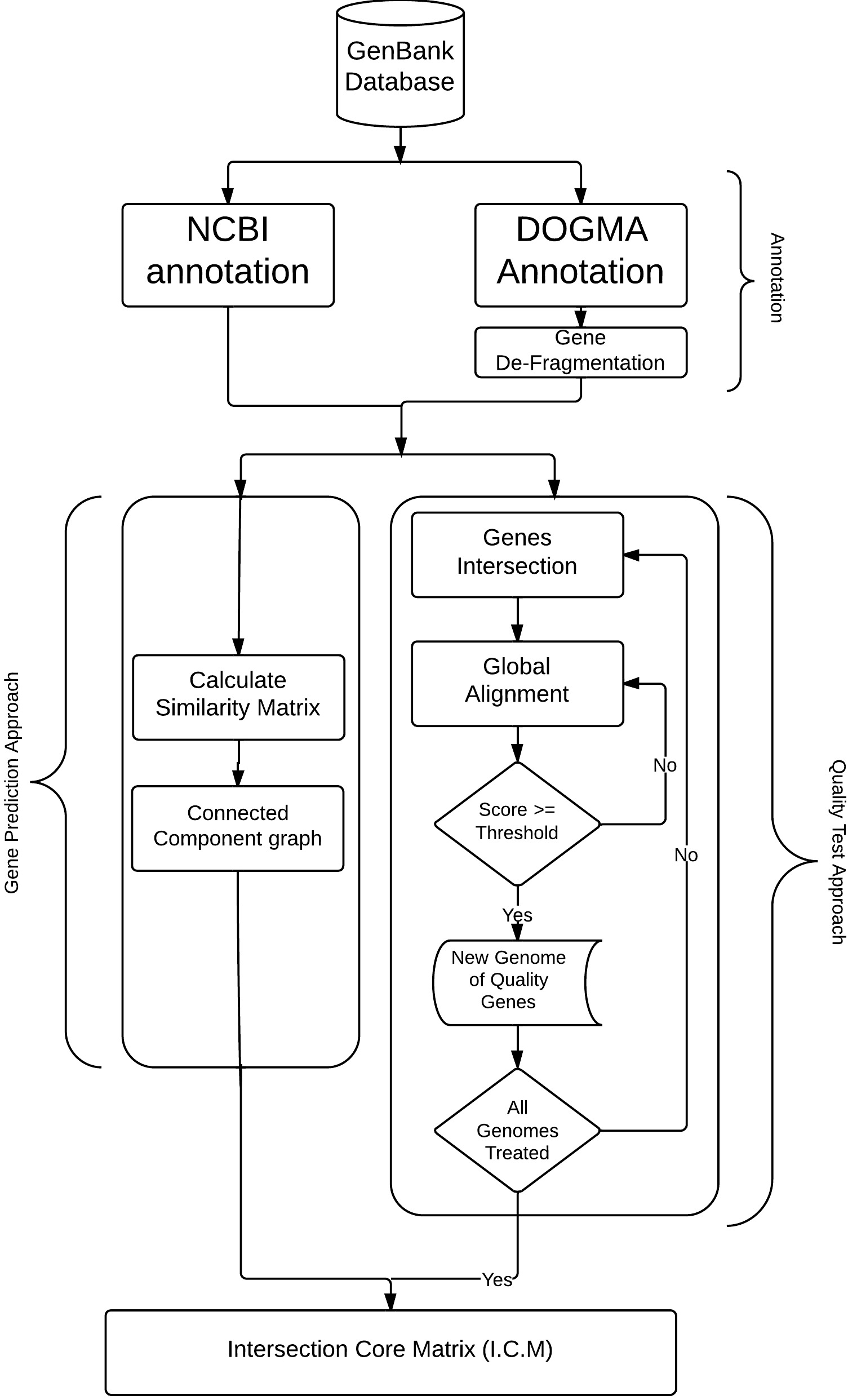}
\caption{An overview of the pipeline}\label{Fig1}
\end{figure}

Instead of considering only gene sequences taken from NCBI or DOGMA \cite{RDogma}, an improved quality test process now takes place as shown in Figure~\ref{Fig1}. It works with gene names and sequences, to produce what we call ``quality genes''. Remark that
such a simple general idea is not so easy to realize, and that it is not sufficient to only consider gene names provided by such tools. 
Providing good annotations is  an important stage for extracting gene features. Indeed, gene features here could be considered as: gene names, gene sequences, protein  sequences, and so on. 
We will subsequently propose  methods that use gene names and sequences for extracting core  genes and  producing  chloroplast evolutionary tree.
 
Real genomes were used in this study, which cover  eleven types  of chloroplast families (see~\cite{Alkindy2014} for more details). 
Furthermore, two kinds of annotations will be considered in this document, namely the
ones provided by NCBI on the one hand, and the ones by DOGMA on the other hand.

\section{Core genes extraction}\label{sec:extraction}
To make this document self contained, we recall the same definition with a fast revision of similarity based method.
    \subsection{Similarity-based approach}\label{sec:simil}
    Basically, this method starts with annotated genomes either from NCBI or DOGMA and uses a distance 
$d:N=\{A,T,C,G\}^{\ast}\times   \{A,T,C,G\}^{\ast}\rightarrow[0,1]$
on genes coding sequences to group similar alleles  in a same cluster.  

For  a given threshold $T\in [0, 1]$  and a similarity measure $d$, 
the method builds  the \emph{similarity} undirected graph  where vertices 
are alleles and s.t. 
there is  an edge between $g_{i}$ and $g_{j}$
if we have $d(g_i,g_j)\leqslant T$.

Each connected component (CC) of this graph defines a class of the 
DNA sequences and is abusively called  a ``gene'',   whereas all its nodes 
(DNA~sequences)  are the ``alleles''  of this  gene. 
Let $\pi$ the function that maps each sequence 
into its representative gene.
Each    genome    
$G=\left\{g_{1}^G,...,g_{m_G}^G\right\}$ is thus mapped into the 
set $\left\{\pi(g_{1}^G),...,\pi(g_{m_G}^G)\right\}$ where  
duplicated genes are removed.

Consequently, the core  genome (resp.,  the pan genome)  of two genomes
$G_{1}$  and $G_{2}$  is defined  as  the intersection  (resp., as  the
union) of their projected  genomes.  The intersection (resp. the union)
of  all the  projected genomes constitutes the core genome 
(resp. the pan genome) of the whole species. 

Let us now consider the 99 chloroplastic genomes introduced earlier. We use in this case study either the coding sequences downloaded from NCBI website or the sequences predicted by DOGMA.
Each genome is thus constituted by a list of coding sequences. In this illustration study, we have evaluated the similarity between two sequences by using a global alignment. More precisely, the measure $d$ introduced in the first approach 
is the similarity score provided after a Needleman-Wunch global alignment, by the \emph{emboss} package released by EMBL~\cite{Rice2000}. 

The number of genes in the core genome and in the pan genome have been computed. 
Obtained results from various threshold values are 
represented in  Table~\ref{Fig:sim:core:pan}. 
Remark that when the threshold is large, the pan genome is large too. 
No matter the chosen annotation tool, this first approach  suffers from producing
too small core genomes,  for any chosen similarity threshold, compared
to   what  is   usually   expected  by   biologists.

\begin{table}[H]
\centering
\caption{Size of core and pan genomes w.r.t. the similarity threshold}
\begin{tabular}{cccccccc}
\hline
          & \multicolumn{4}{c}{Method 1}                          \\ \hline
          & \multicolumn{2}{c}{NCBI}  & \multicolumn{2}{c}{DOGMA} \\ \hline
Threshold(\%) & core       & pan          & core       & pan      \\
\hline
50        & 1          & 163          & 1          & 118          \\
55        & \textbf{5} & \textbf{692} & 2          & 409         \\
60        & 2          & 1032         & 2          & 519          \\
65        & 1          & 1454         & 2          & 685          \\
70        & 0          & 2000         & 1          & 1116        \\ 
75        & 0          & 2667         & 1          & 1781        \\  
80        & 0          & 3541         & 0          & 2730        \\
85        & 0          & 4620         & 0          & 3945        \\
90        & 0          & 5703         & 0          & 5181        \\ 
95        & 0          & 7307         & 0          & 7302        \\
100       & 0          & 8911         & 0          & 10132       \\
\hline
\end{tabular}
\label{Fig:sim:core:pan}
\end{table}
    \subsection{Quality test approach}\label{sec:mixed}
    
\begin{figure}[ht]
\centering
\includegraphics[scale=0.37]{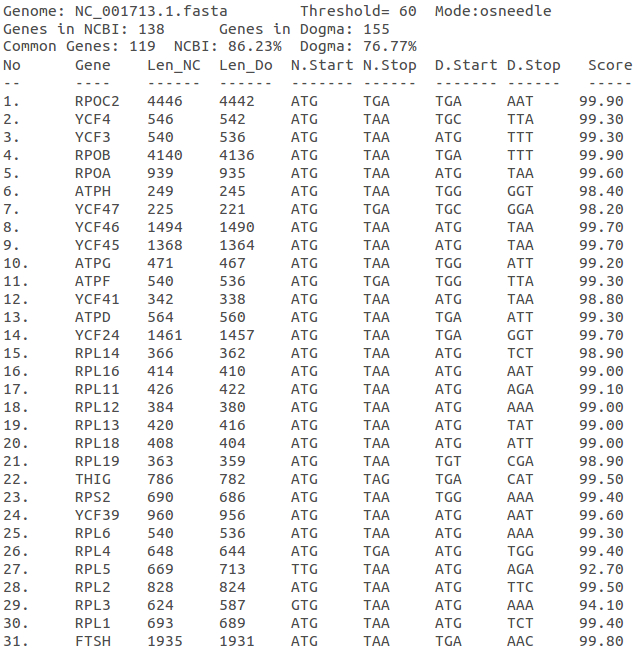}
\caption{Part of the implementation of the second method, comparison of the common genes from NCBI and DOGMA.}
\label{Meth2:gensim}
\end{figure}

Let us present our new approach. In this one, we propose to integrate a similarity distance on gene names into the pipeline. 
Each similarity is computed between a name from DOGMA and a name from NCBI, as shown in Figure~\ref{Meth2:gensim}. 

The proposed distance  is the Levenshtein one, which is close to the Needleman-Wunsch, except that gap opening and extension penalties are equal. The same name is then set to sequences whose NCBI names are close according to this edit distance.
The risk is now to merge genes that are different but whose names are similar (for instance, ND4 and ND4L are two different mitochondrial genes, but with similar names). To fix such a flaw,
the sequence similarity, for intersected genes in a genome, is compared too in a second stage (with a Needleman-Wunsch global alignment) after selecting a genome accession number, and the genes correspondence
is simply ignored if this similarity is below a predefined threshold. We call this operation, which will result in a set of quality genes, a quality test. A result from this quality test process is a set of quality genes. These genes will then constitute the quality genomes. A list of generated quality genomes based on specific threshold will construct the intersection core matrix to generate the core genes, core tree, and phylogenetic tree after choosing an appropriate outgroup.  

It is important to note that DNA sequence annotation raises a problem in the case of DOGMA: contrary to what happens with gene features in NCBI, genes predicted by DOGMA annotation may be fragmented in several parts. 
Such genes are stored in the Gene-Vision file format produced by DOGMA, as each fragment is in this file with the same gene name. A gene whose name is present at least twice in the file is either a duplicated gene or a fragmented one. 
Obviously, fragmented genes must be defragmented before  the DNA similarity computation stage (remark that such a defragmentation has already been realized on NCBI website). As the orientation of each fragment is given in the Gene-Vision output, this defragmentation consists in concatenating all the possible permutations (in the case of duplication), and only keeping the permutation with the best similarity score in comparisons with other sequences having the same gene name, if this score is larger than the given threshold.

\section{Implementation}\label{sec:implem}
All  algorithms have  been implemented using  Python language version 2.7, on a personal computer running Ubuntu~12.04 32bit with 6~GByte memory,
and a quad-core Intel  core~i5~processor with an operating frequency of 2.5~GHz. 

\subsection{Construction of quality genomes}

\begin{figure}[!ht]
    \subfloat[Sizes of genomes based on NCBI and DOGMA annotations. We can see in this figure that the number of genes with DOGMA is larger than with NCBI, because the former generates more tRNAs and rRNAs genes than NCBI.\label{subfig-1:ncbi_vs_dogma}]{%
    \includegraphics[width=0.5\textwidth]{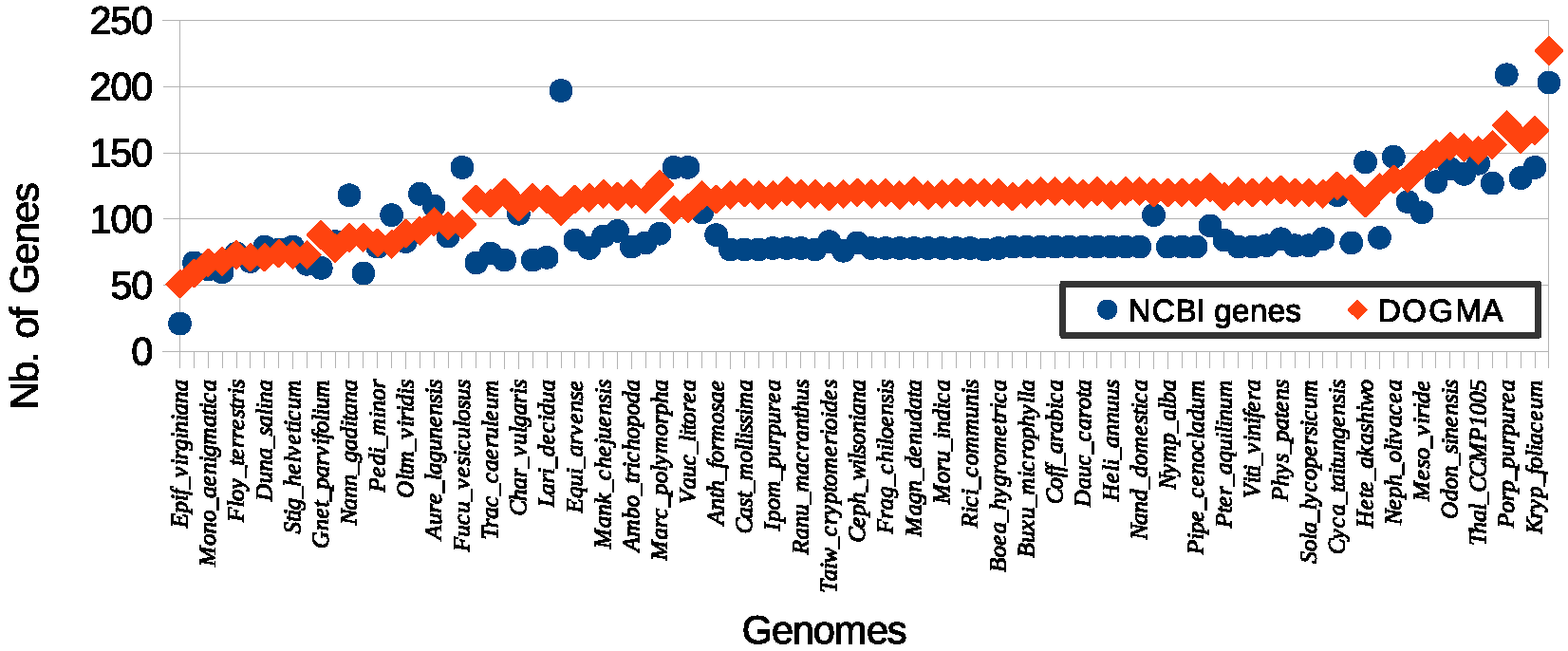}
    }
    \hfill
    \subfloat[Percentage of genes coverage between NCBI and DOGMA. The former outperforms the latter, as almost all genes in NCBI genomes have been covered with common genes, while most of DOGMA genes are ignored. However, correlation of them with NCBI (after quality test) is 0.6731, while it is 0.9664 with DOGMA, this latter being thus more accurate than NCBI. \label{subfig-2:coverage_ncbi_vs_dogma}]{%
      \includegraphics[width=0.5\textwidth]{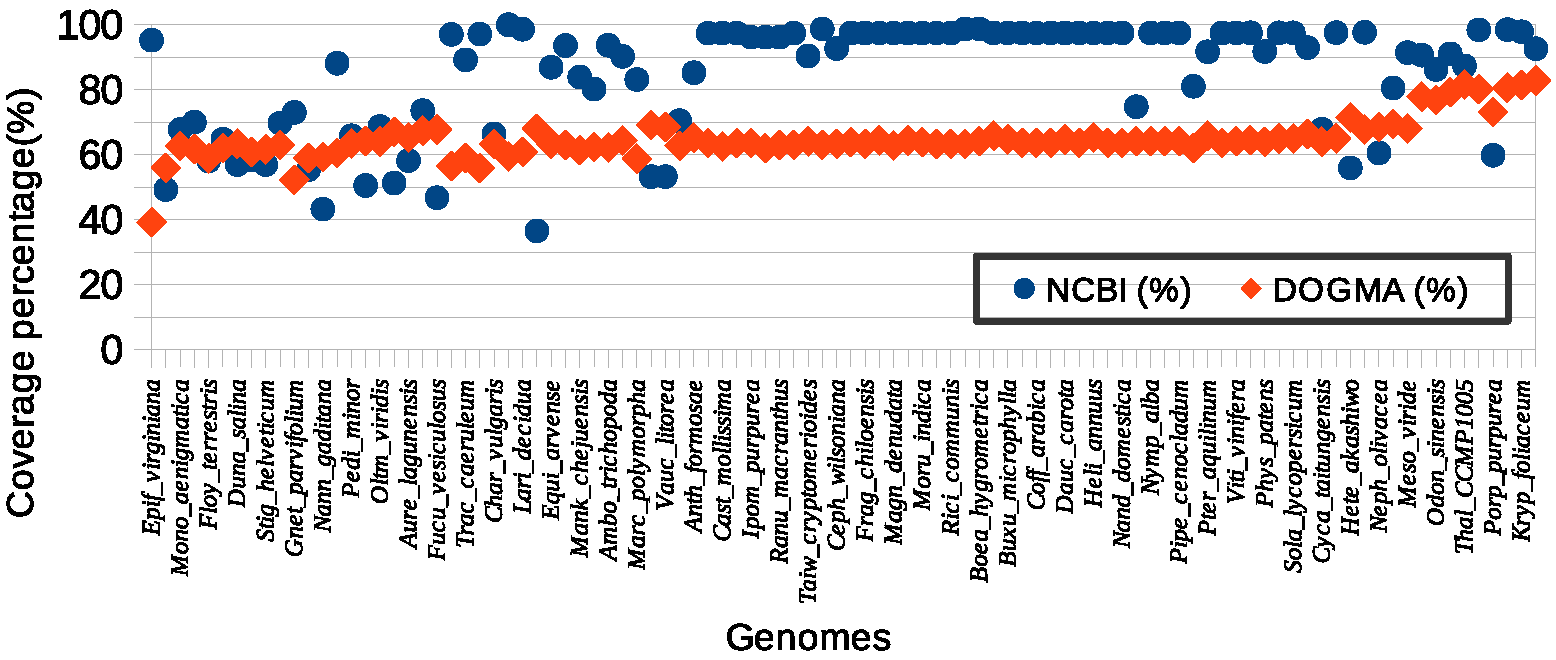}
    }
    \caption{Original and coverage sizes between NCBI and DOGMA genomes based on threshold of 60\%}
    \label{fig:sizes_ncbi_vs_dogma}
\end{figure}

To produce a core tree and genomes based on quality control approach\footnote{see \url{http://members.femto-st.fr/christophe-guyeux/en/chloroplasts}}, we need to know what are the common genes that share almost the same name and sequence from different annotation tools. 
Figure~\ref{subfig-1:ncbi_vs_dogma} shows the original amount of genes based on two different annotation tools, their correlation is equal to 0.57. A two steps quality test routine is then launched
to produce ``quality genomes'' and to enlarge the correlation: (1) select all common genes based on gene names 
and (2) check the similarity of sequences, which must be larger than a predefined threshold. 
Figure~\ref{subfig-2:coverage_ncbi_vs_dogma} presents 
the genes coverage percentage between NCBI and DOGMA. 
Remark that, gene differences between such annotation tools can affect 
the final core genome. More precisely, The number of \emph{tRNAs} and \emph{rRNAs} genes are very high in the case of DOGMA annotation, while they are very low in the case of NCBI. There are also some unnamed or badly named \emph{ORFs} genes in the case of NCBI. 
These genes may improve the final core genome, if their functionality are well defined.  

\subsection{Core and pan genomes}
\begin{figure}[!ht]
    \includegraphics[width=0.49\textwidth]{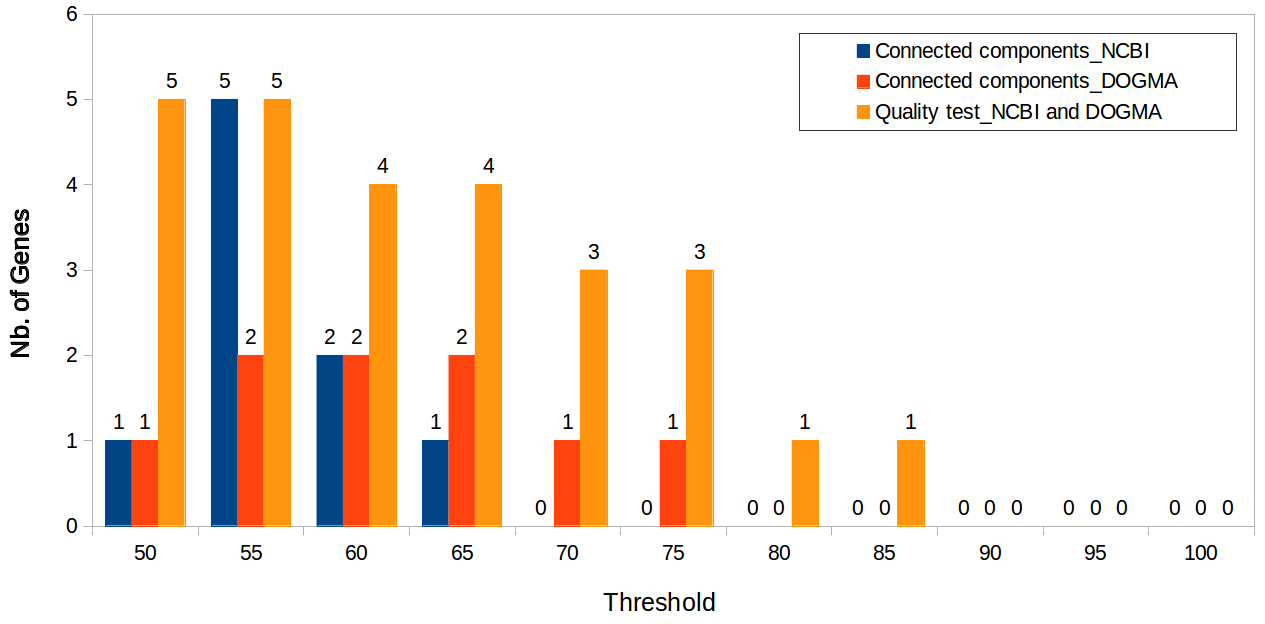}
    \caption{Amount of core genes from each method w.r.t threshold. Note that a maximal number of core genes does not mean good core genomes: we are looking for genes meeting biological requirements.}\label{subfig-1:core}
\end{figure}

The number of {\it  Core genes}, illustrated in Figure~\ref{subfig-1:core}, represents the  amount of genes in  the computed core
genome. The main goal is to  find the largest number of core genes that is compatible
with biological background related to chloroplasts. From the first approach with a threshold of 60\%, we have obtained 2 genes for
99   genomes with NCBI and DOGMA,   whereas 4 genes   for   98   genomes  have been found using
the second approach. In the case of second approach, we have ignored one genome for \emph{Micromonas pusilla} under the accession (NC\_012568.1) from our sample, because we have a few amount of quality genes or none that could have been generated from its correspondents. With the second approach, zero gene in rooted core genome means that we have two or more subtrees of organisms that are completely divergent among each other.
Unfortunately, for the first approach with NCBI annotation, the core genes within NCBI
cores tree did not provide true biologically distribution of the genomes. Conversely, in the case of
DOGMA annotation, the  distribution of genomes is biologically relevant. 
The NCBI under performance may be explained by broken subcores
due to an artificially low  number  of  genes in some genomes intersection, which could be explained
by coding sequence prediction or annotation errors, or by very divergent genomes. 
More precisely, \textit{Micromonas pusilla} (accession number NC\_012568.1) 
is the only genome who totally destroys the final core genome with NCBI
annotations, for both gene features and gene quality methods.

\section{Discussion}
According to chloroplast endosymbiotic theory, the primary endosymbiosis has led to three chloroplast lineages among which the two most evolved groups are the chloroplastida and the rhodophyce{\ae}. These chloroplast groups, which respectively consist of \textit{Land plants} and \textit{Green algae}, and \textit{Red algae}, gave rise to secondary plastids when algae cells were engulfed by other heterotrophic eukaryotes through various secondary endosymbioses. Thus 
\textit{Euglens}~\cite{mcfadden2001primary} come from \textit{Green algae} while 
\textit{Red algae} gave birth to both \textit{Brown algae} and \textit{Dinoflagellates}.

Now, if we observe the built core trees, in particular the one gained with quality control approach, we can notice that a primary plastid generated by the first endosymbiosis can be found in a single lineage of the chloroplast genome evolution tree: the chloroplastida group corresponds to a lineage, whereas the rhodophyce{\ae} group is represented by a second one. The generated core tree is composed by two subtrees, the first one containing the lineages of land plants and green algae and the second one presenting the lineages of brown and green algae. In the tree, some chloroplast lineages such as \textit{Angiosperms} and \textit{green algae} have well biological distributions, while other lineages (\textit{Euglens}, \textit{Dinoflagellates}, and \textit{Ferns}) are badly distributed when compared to their biological history. 
Indeed, common quality genes from quality control approach are well covered by most NCBI genomes, while a large number of \textit{tRNAs} and \textit{rRNAs} from DOGMA genomes have been lost.

\section{Conclusion}\label{sec:concl}
In this research work, we studied two 
methodologies for extracting core genes from a large set of chloroplastic genomes, and we developed 
Python programs to evaluate them in practice. 
A two stage similarity measure, on names and sequences, is thus proposed for DNA sequences clustering in genes, 
which merges best results provided by NCBI and DOGMA. Results obtained
with this ``quality control test'' are deeply compared with our previous research work, on both
computational and biological aspects, considering a set of 99 chloroplastic genomes. 
Core trees have finally been generated for each method, to investigate 
the distribution of chloroplasts and core genomes. The tree from 
DOGMA annotation has revealed the best distribution of
 chloroplasts regarding their evolutionary history. In particular, it appears to
us that each endosymbiosis event is well branched in the DOGMA core tree.

\bigskip

\bibliographystyle{plain}

\end{document}